\documentclass[11pt, a4paper, logo, copyright]{googlecloud}

\pdftrailerid{redacted}

\makeatletter
\renewcommand\bibentry[1]{\nocitep{#1}{\frenchspacing\@nameuse{BR@r@#1\@extra@b@citeb}}}
\makeatother

\usepackage{kantlipsum, lipsum}
\usepackage{dsfont}

\usepackage[authoryear, sort&compress, round]{natbib}

\usepackage[utf8]{inputenc} %
\usepackage[T1]{fontenc}    %
\usepackage{hyperref}       %
\usepackage{url}            %
\usepackage{booktabs}       %
\usepackage{amsfonts}       %
\usepackage{nicefrac}       %
\usepackage{microtype}      %
\usepackage{wrapfig} %
\usepackage{graphicx} %
\usepackage{soul} %
\usepackage{multirow}
\usepackage{arydshln}
\usepackage{subcaption}
\usepackage{enumitem}
\usepackage{amssymb,amsmath}
\usepackage{tikz, adjustbox}
\usepackage[most]{tcolorbox}
\usepackage{float}
\usepackage{algorithm}
\usepackage{algorithmic}
\usepackage[capitalize,noabbrev]{cleveref}
\usepackage{fancyvrb}

\title{\textsc{Magnet}: Multi-turn Tool-use Data Synthesis and Distillation via Graph Translation}

\correspondingauthor{fanyin20@cs.ucla.edu, hamidpalangi@google.com}

\renewcommand{\today}{}

\author[1 2 *]{Fan Yin}
\author[1]{Zifeng Wang}
\author[1]{I-Hung Hsu}
\author[1]{Jun Yan}
\author[1]{Ke Jiang}
\author[1]{Yanfei Chen}
\author[1]{Jindong Gu}
\author[1]{Long T. Le}
\author[2]{Kai-Wei Chang}
\author[1]{Chen-Yu Lee}
\author[1 * $^\clubsuit$]{Hamid Palangi}
\author[1 $^\clubsuit$]{Tomas Pfister}

\affil[1]{Google}
\affil[2]{University of California, Los Angeles}

\begin{abstract}
Large language models (LLMs) have exhibited the ability to effectively utilize external tools to address user queries. However, their performance may be limited in complex, multi-turn interactions involving users and multiple tools. To address this, we propose \textsc{Magnet}, a principled framework for synthesizing high-quality training trajectories to enhance the function calling capability of large language model agents in multi-turn conversations with humans. The framework is based on automatic and iterative translations from a function signature path to a sequence of queries and executable function calls. We model the complicated function interactions in multi-turn cases with graph and design novel node operations to build reliable signature paths. Motivated by context distillation, when guiding the generation of positive and negative trajectories using a teacher model, we provide reference function call sequences as positive hints in context and contrastive, incorrect function calls as negative hints. Experiments show that training with the positive trajectories with supervised fine-tuning and preference optimization against negative trajectories, our 14B model, \textsc{Magnet}-14B-mDPO, obtains 68.01 on BFCL-v3 and 73.30 on ToolQuery, surpassing the performance of the teacher model Gemini-1.5-pro-002 by a large margin in function calling.

\end{abstract}

\begin{document}

\maketitle

\section{Introduction}
Autonomous agents based on large language models (LLMs) have made remarkable progress on fulfilling complex agentic tasks~\citep{yin2024agent, ma2024agentboard, zhang2024xlam}, benefiting from the high capacity of reasoning and planning of LLMs~\citep{achiam2023gpt, team2024gemini, hui2024qwen2}. Among the skillset for agents, the ability to leverage external tools or application programming interfaces (APIs)~\footnote{The terms, \textit{function calling} and \textit{tool-use}, \textit{function} and \textit{API}, are used interchangeably in this paper.} and interact with humans to perform actions in environments is in the central of successful completion of many agentic tasks. Towards this end, recent LLMs have been tailored for function calling (FC) abilities~\citep{schick2023toolformer, patil2023gorilla, dubey2024llama, yang2024qwen2}, achieving improved performance on benchmarks that simulate real-world APIs~\citep{berkeley-function-calling-leaderboard, yao2024tau, guo2024stabletoolbench, ma2024agentboard}.
\begin{figure*}
    \centering
    \includegraphics[width=0.94\linewidth]{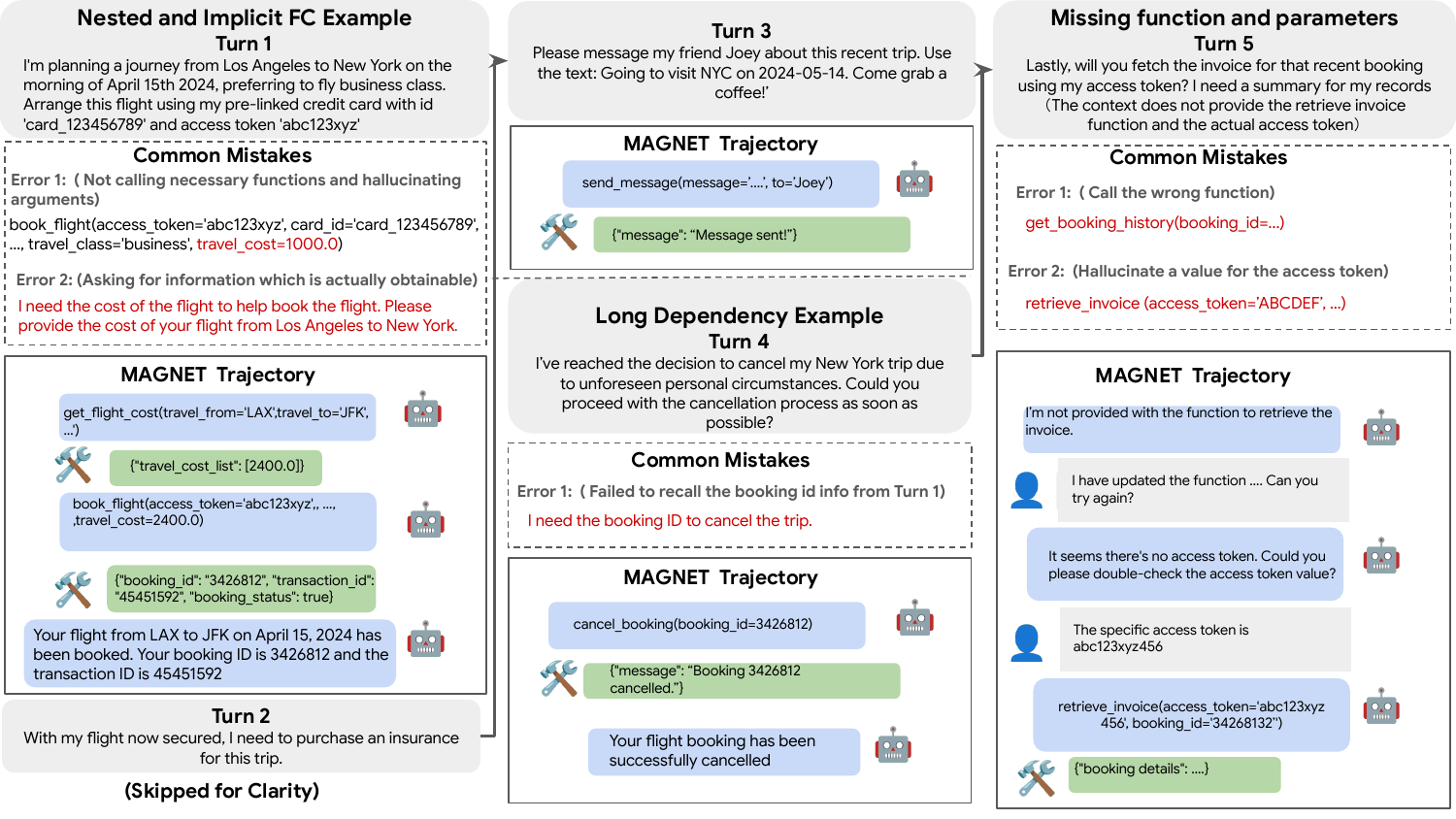}
    \caption{Illustration of challenges and common mistakes in multi-turn FC. An agent needs to understand function outputs and finish follow-up queries from users. This brings several challenges to the agent such as nested FCs (turn 1), long output dependencies (turn 4), irrelevant functions (turn 5). }
    \label{fig:example}
\end{figure*}

However, by qualitatively scrutinizing the behaviors of models, we find that despite the advancements in composing independent FCs, it is still challenging for current LLM agents to perform multi-step and multi-turn interactions with users~\footnote{Multi-step interactions require the LLMs to execute multiple internal FCs to address a single user request, while multi-turn interactions involve an extended exchange between the user and the agents, resulting in multiple conversational turns.} where LLM agents reason, compose FCs and analyze outputs from FCs to respond ~\citep{yao2024tau, berkeley-function-calling-leaderboard}. We summarize three main challenges and common mistakes in multi-turn FC, as illustrated in Figure~\ref{fig:example}: 1) \textit{Nested FCs}: some turns require multiple or even nested FCs which might not be explicitly requested in the query; 2) \textit{Long dependency}: some turns require information from the conversation history to compose FCs; 3) \textit{Irrelevance}: some turns might contain missing functionality or parameter values, for which additional clarification questions are required. Performance-wise, in the Berkeley Function Calling Leaderboard (BFCL-v3)~\citep{berkeley-function-calling-leaderboard}, the best proprietary model achieves 47.62\% success rate on multi-turn cases, while some public models have only around 10\% success rate.

Synthesizing or distilling data from stronger LLMs has been proven a powerful way to improve the reasoning abilities~\citep{guo2025deepseek} of weaker LLMs. Yet the limited performance of existing models on multi-turn cases aggravates the difficulty in gathering high-quality training trajectories to improve the multi-turn ability of public models. To bridge the gap between single FC and multi-turn interactions and build reliable trajectories, we propose a principled pipeline, called \textsc{M}ulti-turn function-c\textsc{A}lling data synthesis with \textsc{G}raph \textsc{T}ranslation, or \textsc{Magnet}, to collect trajectories, i.e., a sequence of user queries, model responses, and tool outputs, to train public models with both supervised fine-tuning (SFT) and preference optimization. 

Our method is based on iterative \textit{back-and-forth translation} (Section~\ref{method:back-forth}). Given a sequence of function signatures, i.e., function names and documentations, we prompt LLMs to iteratively translate them into queries, mimicking user requests, and then compose executable FCs as references. However, forming the \textit{function signature path} (FSP) is not straightforward. Previous works~\citep{qintoolllm} focus on single-turn FCs and randomly sample functions from the same domains. We propose a graph-based approach to constructing multi-turn FSPs.

Motivated by the fact that two functions from the same domain are likely to be relevant in terms of their inputs and outputs, we organize functions as nodes in a graph structure and set a directed edge between two nodes when the source node's outputs relate the target node's inputs. We call them \textit{local dependency graph} as the edges reflect the dependencies among functions. Based on the local dependency graph, we random walk to sample related function signatures and form a FSP. 


From the graph perspective, we find that those challenges mentioned in Figure~\ref{fig:example} can be abstracted as node operations. For example, nested FCs can be abstracted as \texttt{Insert}, which adds extra nodes before another node. Therefore, we further design three node operations: \texttt{Insert}, \texttt{Merge}, \texttt{Split} to enhance the initial FSPs and tailor them to cover the challenges. We show through qualitative (Figure~\ref{fig:example}) and ablation study (Section~\ref{tab:bfcl_ablation}) that including those operations largely improve the reasoning process and reduce common mistakes in multi-turn challenges.

Given the queries and FC references pairs as additional signals, we further control the trajectory generation process with Gemini-1.5-pro-002 as the teacher LLM using context distillation~\citep{snell2022learning}. Specifically, we add FC references as hints while synthesizing trajectories to ensure the quality of positive trajectories. To enable preference-based optimization, we also construct negative trajectories by selecting actions that are making mistakes and deliberately add wrong hints with those mistakes into the trajectories. This makes clear contrary between positive and negative trajectories.

Experiments on two common benchmarks, BFCL-v3 and ToolQuery, demonstrate the advantage of our pipeline. By SFT Qwen2.5-Coder models on 34K trajectories and on 4,556 trajectory pairs with multi-turn direct preference optimization (mDPO)~\citep{xiong2024building}, our model \textsc{Magnet}-14B-mDPO achieves rank 4th on the BFCL-v3 benchmark, surpassing the teacher model Gemini-1.5-pro-002 and adding to the base model by 32.5 points on multi-turn cases. Ablation study shows that all the components in the pipeline is helpful to improving its capability, and the performance improvement can generalize to different base models. Lastly, we show that the trajectory synthesis process can be generalized to other teacher models and self-improve with the base model itself as the teacher model.

Our contributions can be summarized as follows:
\begin{itemize}[itemsep=0.5mm, parsep=0pt, leftmargin=*]
    \item A graph-based perspective for constructing high-quality multi-turn queries and FC references, covering the challenges in multi-turn FC.
    \item A novel technique to distill the information provided in FC references to construct training trajectories for both SFT and mDPO.
    \item We demonstrate superior performance on BFCL-v3 and ToolQuery benchmarks with public models trained with our data. Detailed ablation study shows the effectiveness of each component.
\end{itemize}

\section{Related Work}
\noindent {\bf FC agents evaluation} The ability to use external tools to solve a complex task when the agent lacks some knowledge intrinsically is crucial in agentic behaviors. A variety of benchmarks have been constructed to evaluate such ability. We roughly categorize them as follows based on the amount of functions needed for each test instance and the interactions among functions: (1) single-step; (2) multi-steps, which can be further decomposed into parallel, multiple (chained but not nested), nested; (3) multi-turns. Among those, BFCL-v3~\citep{berkeley-function-calling-leaderboard} is a comprehensive benchmark evaluating single-step, multi-steps, multi-turns scenarios. NexanRaven~\footnote{\url{https://nexusflow.ai/blogs/ravenv2}}, Toolbench~\citep{qintoolllm}, StableToolbench~\citep{guo2024stabletoolbench} mainly test for multi-steps tool-use.~\citet{basu2024nestful} target nested API calls.~\citet{yao2024tau, ma2024agentboard, lu2024toolsandbox} feature multi-turns and multi-steps FCs. However, most of the above mentioned datasets are human curated (with the assistant of LLMs). In contrast, our framework requires minimal human efforts.

\noindent {\bf Training FC agents} Due to the lack of training trajectories, fine-tuning a tool-use agent typically starts with collecting training data. Toolformer~\citep{schick2023toolformer} replaces segments in texts with API calls to train LLMs.~\citep{qintoolllm, chen2024re} synthesize queries from random sampled APIs without clear structure. The xLAM and APIGen series~\citep{liuapigen, zhang2024xlam} unify the format of tool-use data with other agentic tasks and automatically generate queries from verified APIs.~\citet{lin2024hammer} improves the APIGen~\citep{liuapigen} dataset and propose to add function masking and more irrelevant functions to improve the robustness of agents.~\citet{abdelaziz2024granite} introduce fine-tuning with multi-task (function calling, instruction tuning) on 110k data.~\citet{liu2024toolace} synthesize new APIs automatically and directly prompts LLMs to role-play users, agents, and tools.~\citet{chen2024facilitating} adapt composition to improve the quality of single-turn function calling. Among those works,~\citet{qintoolllm, chen2024re, liuapigen} back-translate queries from APIs. While our query generation technique adopt similar ideas, to adapt to multi-turn cases, we propose to organize function signatures in graphs and apply node operations to improve graph complexity. Our trajectory synthesis methods also diverge from previous methods by incorporating more controls. 
\section{Methodology: \textsc{Magnet}}
\begin{figure*}
    \centering
    \includegraphics[width=0.95\linewidth]{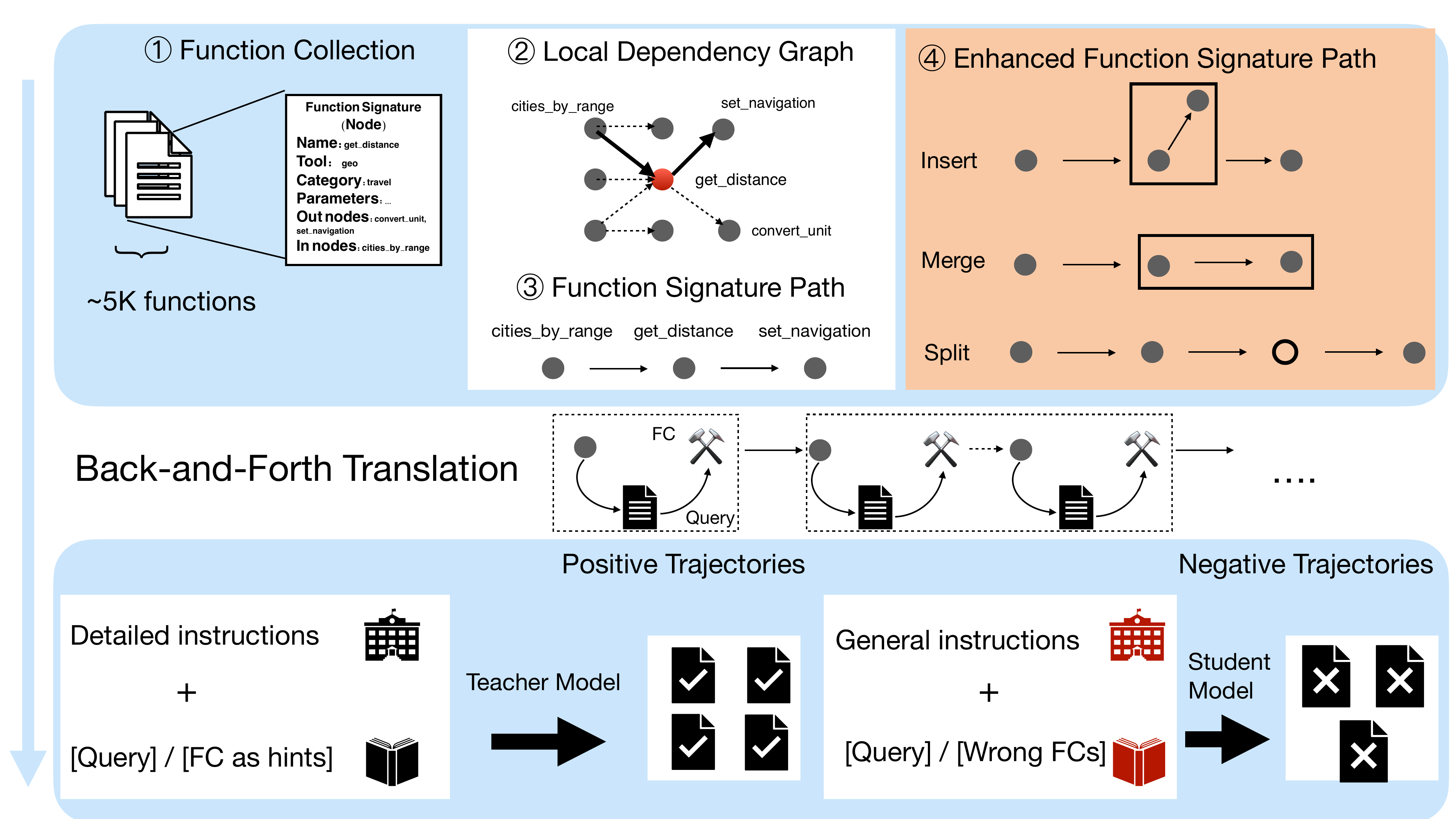}
    \caption{The pipeline for constructing trajectories of function calling. We divide the pipeline into four parts and depicts each part respectively. (1) Construction of the function pool and function execution graph; (2) Node operations defined on the function execution graph; (3) Back-and-forth translation to iteratively create multi-turn queries and fill in function parameters; (4) Construction of positive and negative trajectories by context distillation of good and bad hints and instructions.}
    \label{fig:pipeline}
\end{figure*}
In this section, we first discuss the whole training pipeline to provide more context (Section~\ref{sec:training}). Then, we dive into our main contribution of synthesizing high-quality FC trajectories (Section~\ref{sec:data}).

\subsection{Training setup and formulation}
\label{sec:training}
We leverage SFT and RLHF training. In the first stage, suppose we have a base model and a set of training trajectories $\left\{ \tau^i \right\}$, $i=1\,\dots\,\ n$. Each trajectory involves a sequence ($H$-turns) of user queries, model actions, and tool responses: $\tau^i_w = \left(q^i_1, a^i_1, t^i_1\,\ \cdots\,q^i_H, a^i_H, t^i_H\right)$. The SFT training uses maximum likelihood estimation (MLE) to fit the model actions $a^i_1 \cdots a^i_H$, i.e., the blue parts in Figure~\ref{fig:example}, given the rest as context. Then, in the second stage, given the SFT model and trajectory pairs $\left\{ \tau^i_w, \tau^i_l \right\}$ , $i=1\,\dots\,\ m$, we adopt the mDPO loss~\citep{xiong2024building} plus a MLE loss to further tune the SFT model:
\resizebox{1.05\linewidth}{!}{
 \begin{minipage}{\linewidth}
\begin{align*}
    &\mathcal{L} \left(x; \tau_w, \tau_l\right) =\ \mathcal{L}_\mathrm{SFT} \left(x; \tau_w\right)  + \lambda \mathcal{L}_\mathrm{mDPO} \left(x; \tau_w, \tau_l\right) \notag , \cr
    &\mathcal{L}_\mathrm{mDPO} \left(x; \tau_w, \tau_l\right) =\ \text{-log} \sigma \left(\eta \left(\sum_{\tau_l} \frac{\pi_{\theta}\left(a^{l} | s^{l} \right)}{\pi_{ref}\left(a^{l} | s^{l} \right)} - \sum_{\tau_w} \frac{\pi_{\theta}\left(a^{w} | s^{w} \right)}{\pi_{ref}\left(a^{w} | s^{w} \right)}\right)\right),
\end{align*}
\label{equ:sft}
  \end{minipage}
}
where $\tau_l$ and $\tau_w$ represents the negative and positive trajectories. $\lambda$ is a weight hyperparameter for balancing the two losses, $\pi_{ref}$ is the SFT reference policy, and $\pi_{\theta}$ is the mDPO policy. Next, we dive into \textsc{Magnet}, our method that synthesizes $\left\{ \tau^i \right\}$, $i=1\,\dots\,\ n$ and $\left\{ \tau^i_w, \tau^i_l \right\}$, $i=1\,\dots\,\ m$.

\subsection{Trajectory Construction Overview}
\label{sec:data}
Overall, \textsc{Magnet} first generates pairs of queries and FC references, and then, transform them into trajectories using a teacher LLM. In the first part, the backbone is a back-and-forth translation process inspired by~\citet{nguyen2024better, liself} that converts FSPs into query-reference pairs, which will be introduced in section~\ref{method:back-forth}. The key innovation, however, lies in how we construct high-quality multi-turn FSPs from a graph perspective and node operations (Section~\ref{method:initgraph}). In the second part, we collect both positive and negative trajectories with our newly designed \textit{context distillation} technique. An illustration of the pipeline is in Figure~\ref{fig:pipeline}. For the whole process, we prompt an LLM to help us on tasks like rewriting, back-and-forth translation etc. Without extra statement, we will use Gemini-1.5-pro-002 as the assistant LLM. All the prompts mentioned in this section are in Appendix~\ref{appendix: prompt}.
 
\subsection{Backbone: back-and-forth translation}
\label{method:back-forth}
We start with the back-and-forth translation process, which takes a FSP as input and outputs a sequence of user queries and executable FC references. The same translation process repeats to each function signature. Formally, suppose the whole interaction has $H$ turns, we have the following concepts and notations:

\noindent{\bf Function signature path (FSP):} the function names and attributes (documents, parameters, etc.) for $H$ turns. We denote the FSP as $\mathcal{\phi}\ =\ \left(f_1,\ f_2,\ \cdots\,\ f_H\right)$, where each $f_h$, $h \in \left[H\right]$ denotes function signatures used at turn $h$, which might consists of one of more consecutive functions $f_h = \left(f_{h1}, f_{h2}, \cdots\, f_{hk}\right)$. 

\noindent{\bf Queries:} the sequence of user queries for $H$ turns. We denote them as $\mathcal{Q}\ =\ \left(q_1,\ q_2,\ \cdots\,\ q_H\right)$.

\noindent {\bf Executable functions:} the sequence of executable function call references with actual argument values. We express them in the format of \textsc{func\_name(arg1=val1, ...)} and denote them as $\mathcal{FC}\ =\ \left(fc_1,\ fc_2,\ \cdots\,\ fc_H\right)$. Like function signatures, each turn $h$ contains one or more function calls: $fc_h = \left(fc_{h1}, fc_{h2}, \cdots\, fc_{hk}\right)$. Note that each $fc_{hi}$ maps to a $f_{hi}$. After executing each function call, we will also obtain the corresponding function outputs for each step $i$ at each turn $h$. We denote it as $t_h = \left(t_{h1}, t_{h2}, \cdots\, t_{hk}\right)$.

With the notation above, the back-and-forth translation can be represented as two conjunct functions. For back-translation $\mathcal{M}_b$, it uses the assistant LLM to transform each $f_h$ into a synthetic query $q_h$: 
$$\mathcal{M}_b\left(f_h\right)\ =\ q_h.$$

For the forth-translation $\mathcal{M}_f$, it uses the assistant LLM to transform the $q_h$ into $fc_h$, given $f_h$ and outputs of last round $t_{h-1}$ as additional information, supposing $t_0$ is empty:
$$\mathcal{M}_f\left(q_h, f_h, t_{h-1}\right)\ =\ fc_h.$$

This process is conducted iteratively for each function signature in the FSP to make sure that the outputs from the previous rounds are ready before passing to the later rounds, as illustrated in (3) in Figure~\ref{fig:pipeline}. From the next section, we explain the missing parts from the beginning to the final trajectory, i.e., 1) how we obtain the FSP, $\mathcal{\phi}\ =\ \left(f_1,\ f_2,\ \cdots\,\ f_H\right)$, and 2) how we obtain $\tau = \left(q_1, a_1, t_1\,\ \cdots\,q_H, a_H, t_H\right)$ starting from the queries and executable functions $\left(q_1,\ fc_1,\ t_1,\ \cdots\,q_H,\ fc_H,\ t_H\right)$.


\subsection{Obtain FSP with function dependency graph random walk and node operations.}
\label{method:initgraph}
\paragraph{Function collection and initial FSPs} 
Suppose we have a pool of $N$ functions and their corresponding attributes. We treat functions in the pool as nodes $\{v^1, v^2, \dots, v^{N}\}$ and aim to build a \textit{local dependency graph} $G_i\ =\ \left(\mathcal{N}\left(v^i\right), E_i\right)$ for each target node $v^i$, where $\mathcal{N}\left(v^i\right) = \{c^1, c^2, \dots, c^{|V^i|}\}$ denotes candidate relevant functions or neighbors for $v^i$. $\left(v^i, c^k\right) \in E^i$ denotes edges between $v^i$ and a relevant function $c^k, k \in |V^i|$ which represents they are indeed relevant. 

Specifically, our function collection inherits from previous works~\citep{liuapigen}. We collect the underlined function source codes from the StableToolBench~\citep{guo2024stabletoolbench} and BFCL-v3 multi-turn function implementation~\citep{berkeley-function-calling-leaderboard}. For functions in BFCL-v3, we rewrite the function name and descriptions using our assistant LLM, i.e., only the real implementation that is not exposed to models are kept. For StableToolBench, following APIGen, we select those that contain parameters and are executable verified by simulated calls. In total, we collect 5,011 APIs. For function attributes, besides the function description, arguments information, and response information, we prompt the assistant LLM to label their category and class. For example, a function like \texttt{get\_current\_weather} will have category \texttt{Weather} and the tool class \texttt{Weather condition tool}. Our categories are cleaned based on StableToolBench, which include 49 categories. 

Then for each $v^i, i \in \left[N\right]$, we set $|\mathcal{N}\left(v^i\right)|\ =\ 30$, i.e., we randomly sample 30 candidate nodes from the same category and class as neighbors. Then, we prompt the assistant LLM to judge whether there are dependencies between $v_i$ and $c_k$ based on their inputs and outputs information.

To sample initial FSPs, we conduct a random walk with the local dependency graph $G^i$ for each $v^i$. The random walk starts from $v_i$ and proceed for $S=7$ steps. At each step, we uniformly sample from the out edges of this node and use the target node as the next step, i.e., $\tilde{f}^i_1 = v^i,\ \tilde{f}^i_h \sim \{c^k | {c^k \in \mathcal{N}\left(f^i_{h-1}\right) \text{and} \left(f^i_{h-1}, c^k\right) \in E^{f^i_{h-1}}}\}$ . We denote the initial FSP as $\mathcal{\tilde{\phi}}\ =\ \left(\tilde{f}_1,\ \tilde{f}_2,\ \cdots\,\ \tilde{f}_H\right)$.

\paragraph{Node operations for enhanced FSPs}
To better cover the challenges for multi-turn interactions, we propose to enhance the initial FSPs obtained above with graph-level operations which abstracts the three challenges: nested FCs, long dependency, and missing information.

\noindent{\bf \texttt{Node OP \#1: Insert}} is designed for handling the nested and implicit function call and long dependency scenario. Consider the query: 

\texttt{\small Please check how many kilometers to go from San Francisco to San Mateo},\\ 
which should invoke two functions: 

\texttt{\small get\_distance(from\_loc,to\_loc), convert\_unit(in\_value=<milage obtained from SF\\ to SM>, out\_value)}.

The first function will return a distance in mileage and we need the second function to convert them into kilometers. However, the second function is not mentioned explicitly in the query to be called. Models might not recognize to call the second function. To cover this, our \texttt{Insert} operation will insert an implicit function signature into the current FSP if they are nested. \texttt{Insert} will also be useful for creating examples covering the long dependency challenge. For example, we could add another \texttt{cities\_by\_range} in a few rounds later which reuses the outputs from \texttt{get\_distance}.

Formally, we iterate through the FSP $\mathcal{\tilde{\phi}}\ =\ \left(\tilde{f}_1,\ \tilde{f}_2,\ \cdots\,\ \tilde{f}_H\right)$. For each turn in FSP $\tilde{f}_h = \left(\tilde{f}_{h1}, \dots, \tilde{f}_{hk}\right)$ which consists of one or multiple function signatures, we select the last function signature $\tilde{f}_{hk}$ and use our assistant LLM to check if any of its neighbors $\mathcal{N}\left(\tilde{f}^i_{hk}\right)$ satisfies the requirements for a nested function signature (see the prompts to judge nested functions in Appendix~\ref{appendix: prompt}). If so, we collect that function signature, denote as $c_{hk}$, and append it to the current turn. Finally, we obtain $f_h = \left(\tilde{f}_{h1}, \dots, \tilde{f}_{hk}, c_{hk}\right)$. We may also insert $c_{hk}$ as an individual turn after a random later turn to reflect long dependency.

\noindent{\bf \texttt{Node OP \#2: Merge}} is for creating a single-turn query that would involve multiple function calls and cover short dependency. Notice that the key difference with \texttt{Insert} and nested API calls is that we could \texttt{Merge} multiple functions that are relevant but not exactly nested. In this case, agents should understand the outputs from the previous functions in this turn to compose the consecutive function. For example, the following query would invoke both \texttt{get\_distance(from\_loc,to\_loc)}, \texttt{set\_navigation(distance)}:

\texttt{\small Can you check how many kilometers to go from San Francisco to San Mateo and \\ then set up the navigation for me with the obtained distance?} 

Formally, for the initial FSP $\mathcal{\tilde{\phi}}\ =\ \left(\tilde{f}_1,\ \tilde{f}_2,\ \cdots\,\ \tilde{f}_H\right)$, we take each two consecutive turns $\tilde{f}_h$ and $\tilde{f}_{h + 1}$ and combine them into one turn $f_h = \left(\tilde{f}_h, \tilde{f}_{h + 1}\right)$ with a probability of $p=0.3$. 

\noindent{\bf \texttt{Node OP \#3: Split}} is mainly designed for the missing or irrelevant function information scenarios. For the previous query, if the function \texttt{get\_distance} is not provided, or the query omits the destination: \texttt{\small Please check how many kilometers to go from San Francisco to somewhere}, the agent should ask a clarification question. Formally, we randomly select a turn $\tilde{f}_h$ in the initial FSP and split it into two turns $f_h = \tilde{f}_h$, $f_{h + i + 1} = \tilde{f}_{h + i}\ \text{where}\ i=1,\dots,H - h - 1$, and $f_{h + 1} = \{\}$. The null node will be labeled with `miss params' or `miss func' which will act as an indicator when translating. 

For each FSP, we sequentially apply \texttt{Merge} then \texttt{Insert} to obtain an enhanced FSP $\mathcal{\phi}\ =\ \left(f_1,\ f_2,\ \cdots\,\ f_{H^*}\right)$. We use $H^*$ to denote the new number of turns. Then, we will apply \texttt{Split} on $\mathcal{\phi}$ to obtain another enhanced FSP with missing information $\mathcal{\hat{\phi}}\ =\ \left(f_1,\ f_2,\ \cdots\, f_h, \{\}, \cdots \ f_{H^* + 1}\right)$. Both $\mathcal{\phi}$ and $\mathcal{\hat{\phi}}$ will go to the next step for back-and-forth translation. For consistency and simplicity, we call them enhanced FSP and use the unified $\mathcal{\phi}$ as the notation.

\subsection{Hint-based context distillation for positive and negative trajectory sampling}
\label{method:traj}

Recall that after obtaining enhanced FSPs, we will leverage the back-and-forth translation to convert them into queries $q_h$ and executable FCs $fc_h$. In this step, we transform $\left(q_1,\ fc_1,\ t_1,\ \cdots\ ,\ q_H,\ fc_H,\ t_H\right)$ to $\tau = \left(q_1,\ a_1,\ t_1\,\ \cdots\ ,\ q_H,\ a_H,\ t_H\right)$, which essentially transform from function call references $fc_h$ into model actions $a_h$. The model action $a_h$ for each turn may consist of different components. For example, with the ReAct~\citep{yaoreact} style, the model actions synthesize verbal reasoning, FCs, and textual summarization of function outputs, while with other styles, the explicit verbal reasoning might be omitted. When generating positive trajectories, we hope the teacher model to be as accurate as possible, where the model actions fully cover the functionality of $fc_h$ and compose precise response to users based on the functions However, no current LLMs can consistently produce perfect trajectories.

Inspired by context distillation~\citep{snell2022learning}, we propose to add a \texttt{[Hint]} section after each query $q_h$ to indicate the functions being called during this turn, using the $fc_h$, and provide detailed instructions when sampling the positive trajectories. We show the prompts in Appendix~\ref{appendix: prompt}.

On the contrary, when generating negative trajectories, we hope the trajectories reflect the mistakes made by models. So for negative trajectories, we also include such hints but the actual content is a misleading wrong FCs. Those FCs are collected from the mistakes of the SFT model. Specifically, for each data instance, we collect ten trajectories from the SFT model. Then, for each turn in each trajectory, we present it to the assistant LLM as a judge to decide whether this turn includes an incorrect FC that conforms with any one of the errors defined in the judgement prompt. If so, we will collect them as misleading hints when prompting SFT model to sample negative trajectories.

\subsection{Post-processing and data mixture}
We adopt the following post-processing techniques to enhance the diversity of the SFT datasets so that models trained with our data could be more robust to variations in superficial features.
\begin{itemize}[itemsep=0.5mm, parsep=0pt, leftmargin=*]
    \item For each training data, we shuffle the order of available functions in system prompts.
    \item We filter out trajectories with rule-based metrics: we collect several key words that indicate failed FCs at the end of each turn, such as `Bad request'. `does not match' etc. This roughly excludes incorrect trajectories or wrong formatting in FCs. 
    \item Besides multi-turn data, we add the following types of data into our final SFT data mix: single-turn data including those that invoke single, parallel (same function with different arguments), and multiple (different but relevant) FCs. This is for warming up the model on function calling; 2) irrelevance functions where models should be able to detect. A study on how to mix those data is provided in Section~\ref{exp:ana}.
 
\end{itemize}

\subsection{Data statistics}
\begin{wraptable}[11]{C}{0.44\textwidth}
\footnotesize
\setlength\tabcolsep{3pt}
\scalebox{0.95}{
\begin{tabular}{lccc}
\toprule
\textbf{Category} & \textbf{\# SFT} & \textbf{\# mDPO}\\
\midrule
Single-turn & 20,000 & 1,556 \\
Multi-turn & 7,800 & 2,250 \\
Irrelevance & 6,200 & 750 \\
Avg. FCs (for single-turns) & 1.80 & 1.94 \\
Avg. Turns (for multi-turns) & 4.71  & 5.22 \\
Avg. FCs (for multi-turns) & 15.13 & 14.98 \\
\bottomrule
\end{tabular}
}
\caption{Data statistics for the training sets. \# SFT and \#mDPO represents the number of samples in SFT and mDPO training sets of the corresponding category.}
\label{tab:stats}
\end{wraptable}
Our final SFT training set contains 34,000 instances and the preference learning set contains 4,556 instances. The total training size, 38,556, is around half of other current public datasets such as APIGen (60,000), Hammer (67,500) etc. We present a detailed statistics about the number of each data type, the number of turns, and the number of function calls in Table~\ref{tab:stats}. A study about function contamination is presented in Appendix~\ref{appendix: contamimation}.

\section{Experiments}
\setlength{\tabcolsep}{3.5pt}
\begin{table*}[ht]
 \centering
 \resizebox{0.95\linewidth}{!}{
 \begin{tabular}{c | c | c c c | c c c c c| c c}
 \hline
 \multirow{2}{*}{\textbf{Model}} & \multirow{2}{*}{\textbf{Overall}} & \multicolumn{3}{c|}{\textbf{Single Turn}} & \multicolumn{5}{c|}{\textbf{Multi-turn}} &  \multicolumn{2}{c}{\textbf{Hallucination Measure}}  \\
& & \textbf{Non-live AST} & \textbf{Non-live Exec} & \textbf{Live AST} & \textbf{Overall} & \textbf{Base} & \textbf{Miss Func} & \textbf{Miss Param} & \textbf{Long} & \textbf{Relevant} & \textbf{Irrelevant}\\

 \hline
 \multicolumn{12}{c}{\textbf{Top six models}} \\
 \hdashline
 \textsc{watt-tool-70B (FC)} & 74.31 & 84.06 & 89.39 & 77.74 & 58.75 & 67.50 & 57.50 & 48.50 & 61.50 & 94.44 & 76.32 \\
 \textsc{gpt-4o-2024-11-20 (Prompt)} & 72.08 & 88.10 & 89.38 & 79.83 & 47.62 & 59.00 & 41.00 & 35.50 & 55.00 & 83.33 & 83.76 \\
 \textsc{gpt-4o-2024-11-20 (FC)} & 69.58 & 87.42 &	89.20 &79.65	& 41.00	&62.50	&6.00	&37.50	&58.00  &83.33	&83.15\\
\textsc{GPT-4-turbo-2024-04-09} & 67.88 & 84.73&	85.21&	80.50&	38.12&	54.00&	13.50&	35.50&	49.50& 72.22&	83.81 \\
\textsc{watt-tool-8B* (FC)} &  67.33 & 86.44 &	87.73&	76.23&	38.25 &	46.00&	40.00&	27.00 &	40.00 &	77.78 &	82.89 \\
\textsc{o1-2024-12-17 (Prompt)} & 66.73 & 78.92&	82.70&	78.14&	28.25&	40.50&	5.00&	34.50&	33.00&	61.11&	89.62 \\

 \hline
 \multicolumn{12}{c}{\textbf{Gemini models (teachers)}} \\
 \hdashline
 Gemini-1.5-Pro-002 (Prompt) & 62.19 & 88.58	& 91.27	&76.72	&20.75	&23.00	&19.50	&17.50	&23.00	&72.22	&78.15 \\
Gemini-2.0-Flash-Exp (Prompt) & 61.74 & 89.96&	79.89&	82.01&	17.88&	28.00&	3.00&	19.00&	21.50&	77.78&	86.44 \\
\hline
 \multicolumn{12}{c}{\textbf{7B models}} \\
 \hdashline
 Functionary-Small-v3.1 (FC) & 56.49 & 86.75 & 87.12 & 73.75 & 10.12 & 18.00 & 2.50 & 14.00 & 6.00 & 77.78 & 70.89 \\
  Hammer2.1-7b (FC) & 61.83 & 88.65&	85.48&	75.11&	23.50 &	35.50 &	\bf 25.50&	19.00	&14.00&	82.35& 78.59 \\
 Qwen2.5-Coder-7B-Instruct & 53.13 & 86.83 & 82.27 & 66.99 & 8.25 & 11.50 & 6.50 & 5.50 & 5.50 & \bf 88.89 & 65.39  \\
\textsc{Magnet}-7B-SFT &  62.73 & 88.60 & 85.73 & 74.19 & 26.50 & 35.50 & 24.00 & \bf 27.50 & 19.00 & 66.67 & \bf 78.67 \\
\textsc{Magnet}-7B-mDPO & \bf 64.64 & \bf 89.40 & \bf 89.27 & \bf 77.92 & \bf 27.75 & \bf 39.00 & 24.00 & 26.00 & \bf 22.00 & 83.33 & 78.51 \\
 
\hline
 \multicolumn{12}{c}{\textbf{14B models}} \\
 \hdashline
Qwen2.5-Coder-14B-Instruct & 51.88 & \bf 90.94 & 87.80 & 65.30 & 5.38 & 7.50 & 7.00 & 4.00 & 3.00 & \bf 100.00 & 44.58  \\
MAGNET-14B-SFT & 66.83 & 90.02 & 88.20 & 77.92 & 33.38 & 47.00 & 32.00 & 32.00 & 22.50 & 72.22 & 82.59 \\
\textsc{Magnet}-14B-mDPO & \bf 68.01 & 90.13 & \bf 89.75 & \bf 79.14 & \bf 37.88 & \bf 52.00 & \bf 36.00 & \bf 35.50 & \bf 28.00 & 88.89 & \bf 84.78\\

 \hline
\end{tabular}
 }
 \vspace{-1mm}
 \caption{Main results on BFCL-v3. Our \textsc{Magnet} series demonstrate substantial improvements compared to their base model, Qwen2.5-Coder series, in both the multi-turn function calling and overall evaluations. Our 14B model ranked \#4 in the leaderboard, surpassing o1 and the teacher model Gemini-1.5-pro-002. Best numbers under each test category and base models are \textbf{bold}. * indicates reproduced results with the exact same process as our models.}
 \label{tab:bfcl_main}
\end{table*}
We conduct experiments on the following two benchmarks: BFCL-v3~\citep{berkeley-function-calling-leaderboard} and ToolQuery~\citep{ma2024agentboard}. BFCL-v3 is a comprehensive benchmark designed for different aspects of function calling, including single-turn, multi-step, multi-turn, and irrelevant function calls categories. ToolQuery is part of a broader agent benchmark that test model's ability in composing multi-step and multi-turn function calls in academia, weather, movie areas. BFCL-v3 have in total 4,751 test cases while ToolQuery contains 60 test cases. We use a unified prompt format for both tasks, as shown in Appendix~\ref{appendix: prompt}.

\subsection{Setup}
\label{exp:setup}
We fine-tune Qwen2.5-Coder-7B-instruct and Qwen2.5-Coder-14B-instruct. For the training, we first train with the 34,000 positive trajectories with SFT. We set a peak learning rate of 1e-5 with warm up and linear decay, and a batch size of 64. Then, in the mDPO stage, we do full fine-tuning on the 7B models and set the learning rate to be 5e-7 and batch size 32. For mDPO on 14B model, we conduct LoRA tuning~\citep{hulora} with a learning rate of 1e-6. More details in Appendix~\ref{appendix: training}.

\subsection{Main results on BFCL-v3}
We compare the performance of our trained model with top ranked and related models on the BFCL-v3 benchmark. Results are presented on Table~\ref{tab:bfcl_main}. The performance of our best 14B model ranks 4th on the leaderboard, surpassing the o1 model and on par with GPT-4-Turbo on both the overall performance and the multi-turn performance. We show that with mDPO on targeted loss patterns, the performance on multi-turn scenarios can be boosted compared to SFT only models, with a margin of 2.50\% success rate for the 14B model. Notice that all of our 7B and 14B models, including SFT and mDPO models, outperform the teacher model Gemini-1.5-pro-002 on the multi-turn scenario. This demonstrates that our data synthesis pipeline introduces additional signals and provides better supervision compared to directly distilling from the teacher model.

Finally, comparing with base models and other public models of the same size, our trained model boosts the performance by 18.5 and 30.0 on multi-turn scenarios for the base 7B and 14B Qwen2.5-Coder models, respectively. We also outperforms Hammer2.1-7b (FC), a competitive FC agent model trained from the same base model.

\begin{wraptable}[11]{C}{0.44\textwidth}
\footnotesize
\setlength\tabcolsep{3pt}
\scalebox{0.95}{

  \begin{tabular}{l | c c }
    \hline
    & \textbf{Success rate} & \textbf{Progress rate} \\
    \hline
    Qwen-Coder-7B-instruct & 15.0 & 34.0 \\
    Qwen-Coder-14B-instruct & 51.7 & 68.7 \\
    GPT-4o & 63.3 & \bf 80.1 \\
    Gemini-1.5-pro-002 & 68.3 & 74,6\\
    xLAM-8x22b-r & 68.3 & 75.8 \\
    \textsc{Magnet}-7B-mDPO & 67.7 & 73.4 \\
    \textsc{Magnet}-14B-mDPO & \bf 73.3 & 78.7 \\
    \hline
  \end{tabular}
  }
  \caption{Main results on ToolQuery. Our 14B model achieved the best performance on success rate.}
  \label{tab:toolquery_main}
  \vspace{-3mm}
\end{wraptable}
\setlength{\tabcolsep}{3.5pt}
\begin{table*}[t]
 \centering
 \resizebox{0.95\linewidth}{!}{
 \begin{tabular}{c | c | c c c | c c c c c| c c}
 \hline
 \multirow{2}{*}{\textbf{Model}} & \multirow{2}{*}{\textbf{Overall}} & \multicolumn{3}{c|}{\textbf{Single Turn}} & \multicolumn{5}{c|}{\textbf{Multi-turn}} &  \multicolumn{2}{c}{\textbf{Hallucination Measure}}  \\
& & \textbf{Non-live AST} & \textbf{Non-live Exec} & \textbf{Live AST} & \textbf{Overall} & \textbf{Base} & \textbf{Miss Func} & \textbf{Miss Param} & \textbf{Long} & \textbf{Relevant} & \textbf{Irrelevant}\\

 \hline
 \multicolumn{12}{c}{\textbf{Ablation on pipeline components}} \\
 \hdashline
 init graph & 58.54 & 89.60 & 87.13 & 76.96 & 12.75 & 14.50 & 13.00 & 13.50 & 10.00 & 94.44 & 78.95 \\
 + merge & 60.83 & 89.76 & 87.81 & 76.92 & 20.63 & 26.50 & 18.00 & 19.00 & 19.00 & 77.78 & 76.87 \\
  + merge + insert & 64.39 & 90.89 & 87.91 & 77.37 & 29.25 & 42.00 & 26.50 & 24.50 & 24.00 & 88.89 & 78.90 \\
MAGNET-14B-SFT & 66.83 & 90.02 & 88.20 & 77.92 & 33.38 & 47.00 & 32.00 & 32.00 & 22.50 & 72.22 & 82.59 \\
\hline
MAGNET-14B-SFT & 66.83 & 90.02 & 88.20 & 77.92 & 33.38 & 47.00 & 32.00 & 32.00 & 22.50 & 72.22 & 82.59\\
 - context-distillation-positive & 60.26 & 88.27 & 84.29 & 76.63 & 18.88 & 21.00 & 20.00 & 15.50 & 19.00 & 72.22 & 78.00 \\

\hline
MAGNET-14B-mDPO & 68.01 & 90.13 & 89.75 & 79.14 & 37.88 & 52.00 & 36.00 & 35.50 & 28.00 & 88.89 & 84.78\\
 - context-distillation-negative & 67.35 & 90.34 & 88.96 & 78.84 & 36.25 & 48.50 & 34.50 & 35.00 & 27.00 & 88.89 & 83.79 \\

 \hline
 \multicolumn{12}{c}{\textbf{Comparison between training data source: Qwen-Coder-7B-instruct as base model}} \\
 \hdashline
 MAGNET-7B-SFT &  62.73 & 88.60 & \bf 85.73 & 74.19 & 26.50 & \bf 35.50 & 24.00 & 27.50 & 19.00 & 66.67 & 78.67 \\
 APIGen + ToolAce &  50.30 & 88.85 & 89.59 & 59.04 & 7.13 & 10.50 & 6.50 & 5.50 & 4.50 & 100.00 & 39.17 \\
 APIGen + ToolAce + Irrelevant & 57.24 & 87.44 &  89.54 & 76.99 & 6.25 & 9.00 & 5.50 & 7.00 & 3.50 & 77.78 &  83.79	\\
Hammer2.1-7b (FC) & 61.83 & 88.65&	85.48&	75.11&	23.50&	35.50&	25.50&	19.00	& 14.00 &b82.35 &	78.59 \\
 
\hline
 \multicolumn{12}{c}{\textbf{Effectiveness of MAGNET across different base models}} \\
 \hdashline
 \textsc{Qwen2.5-Coder-instruct} & 50.01 & 86.15 & 82.45 & 64.46 & 4.25 & 6.00 & 6.50 & 3.50 & 1.00 & 100.00 & 51.60 \\
  \textsc{MAGNET-Qwen2.5-Coder-instruct} &  62.73 & 88.60 & 85.73 & 74.19 & 26.50 & \bf 35.50 & 24.00 & 27.50 & 19.00 & 66.67 & 78.67 \\
 \textsc{Qwen2.5-instruct} & 52.58 & 86.83 & 82.27 & 66.99 & 7.25 & 8.50 & 10.00 & 5.50 & 5.00 & 88.89 & 65.39 \\
 \textsc{MAGNET-Qwen2.5-instruct} & 59.84 & 88.12 & 85.48 & 72.86 & 21.12 & 31.00 & 19.00 & 21.00 & 13.50 & 83.33 & 76.67 \\
 \textsc{Mixtral-8x7B-instruct-v0.1} & 36.93 & 47.94 & 51.59 & 57.71 & 0.50 & 1.00 & 0.00 & 0.00 & 1.00 & 38.89 & 75.37 \\
 \textsc{MAGNET-Mixtral} & 58.17 & 88.46 & 80.20 & 68.46 & 19.75 & 24.50 & 22.50 & 19.00 & 13.00 & 94.44 & 66.47 \\
 
  \hline
 \multicolumn{12}{c}{\textbf{Performance with different teacher models, including self-improvement}} \\
 \hdashline
 Gemini-1.5-pro-002 & 66.83 & 90.02 & 88.20 & 77.92 & 33.38 & 47.00 & 32.00 & 32.00 & 22.50 & 72.22 & 82.59\\
 GPT-4o & 65.79 & 89.76 & 87.86 & 77.42 & 30.88 & 44.00 & 29.00 & 27.00 & 23.5 & 72.22 & 82.34 \\
 Qwen2.5-Coder-14B-Instruct (self-improvement) & 64.33 & 90.04 & 87.64 &  77.79 & 27.12 & 38.50 & 24.50 & 27.50 & 18.00 & 72.22 &  80.86	\\

 \hline
\end{tabular}
 }
 \vspace{-1mm}
 \caption{Ablation results on BFCL-v3. We show the effects of ablating out different components in our data synthesis pipeline. We also compare with different base models, different teacher models, and different data sources. Results demonstrate the effectiveness of our training data from different aspects.}
 \label{tab:bfcl_ablation}
 \vspace{-3mm}
\end{table*}
\subsection{Main results on ToolQuery}

Results for ToolQuery are shown in Table~\ref{tab:toolquery_main}. We achieve a success rate of 73.3 on ToolQuery by training Qwen2.5-Coder-14B-instruct on our data, surpassing the performance of a strong proprietary model, GPT-4o, and a much larger public model tuned on the function-calling task, xLAM-8x22b-r. Notice that all the functions from ToolQuery are unseen in the training set. This further demonstrates the generalization ability of our trained models on unseen functions.

\subsection{Ablation Study and Analysis}

We conduct ablation study to answer the questions: (1) how each component in our pipeline affects the overall performance? (2) how our synthetic data is better than other public training datasets? (3) is the effects of the synthetic data consistent among different base models? (4) how the advantage of the framework transfers to different teacher models, including self-improvement. The full results are presented in Table~\ref{tab:bfcl_ablation}. Findings below:

\noindent {\bf Pipeline design} We conduct experiments to see the effects of local dependency graph construction, each node operation, positive trajectories sampled with correct hints, and negative trajectory sampled with wrong hints in the model performance. As shown in the first part (first six rows) in Table~\ref{tab:bfcl_ablation}, we demonstrate that each component is helpful in the final performance of the model. Especially, with the initial local dependency graph, we are able to improve upon the base model by around 8\% on multi-turn success rate. Building upon that, both merge and insert operations boost the multi-turn performance by a large margin, especially on the base multi-turn test cases. Finally, adding split operation directly helps with the missing function, missing parameters, and irrelevance detection scenarios 5.5\%, 7.5\%, and 3.69\%, respectively. We also observe a substantial boost in performance when we distill FC references into positive trajectories compared to directly distilling Gemini-15-pro-002 trajectories from the multi-turn queries. This brings a 14.50\% gain in multi-turn performance. Finally, adding negative trajectories using our context distillation technique brings around 0.5\% improvements compared to randomly sample rejected trajectories from the SFT model.

\noindent {\bf Data sources} To demonstrate the benefits of our constructed data against other public training data, we train the same base model using different sources of open-sourced data. Specifically, amongst the top-performance models, the only open-sourced training datasets are APIGen~\citep{liuapigen} and a subset of ToolAce~\citep{liu2024toolace}. Further, the Hammer2.1-7b model~\citep{lin2024hammer}, although not open-sourcing the full training data, is trained from the same base model with an augmented dataset with irrelevant functions and masking techniques. Therefore, we compare our model with two other models: the same based model trained with a combination of all open-sourced training data, i.e., APIGen and ToolAce, and Hammer2.1-7b (FC). As shown in the second section in Table~\ref{tab:bfcl_ablation}, our \textsc{Magent}-7B-SFT surpasses other open-source data by a large margin, especially in the multi-turn scenario. We outperform Hammer2.1-7B by 3 points and models trained with APIGen and ToolAce data by 20.25. This demonstrates the effectiveness of our training data. 

\noindent {\bf Base model} We analyze on the effects of base model on the final performance. Besides the original Qwen2.5-Coder-instruct series, we compare with Qwen2.5-instruct series, which are trained without additional code data, and Mixtral-8x7B-instruct-v1. We observe that Coder series models, although obtaining slightly weaker performance on multi-turn and irrelevance detection without fine-tuning on our data, have better potential to learn from the training data, which achieves 5.38 better performance on multi-turn cases. Besides, by training comparing Mixtral-8x7B-instruct-v1 and \textsc{Magent}-Mixtral, we demonstrate that the performance boost brought by our data on function calling can be generalized to other models as well.

\noindent {\bf Teacher model} We analyze on the effects of using different teacher models for synthesizing trajectories. Specifically, we use Gemini-1.5-pro-002 (original setting), GPT-4o as teachers and also investigate using the Qwen2.5-Coder-14B-Instruct itself as the teacher model given the hint-based trajectory distillation. We find that Gemini-1.5-pro-002 and GPT-4o as teachers show comparable performance. This reflects that using the hints as context closes the gap between the original performance of teachers on multi-turn scenarios. Furthermore, self-improvement is also possible under the current data generation framework, despite of slight performance drops.

\noindent{\bf Discussion: the impact of data mixture}
\label{exp:ana}

\begin{wrapfigure}[15]{R}{0.5\textwidth}
\footnotesize
\setlength\tabcolsep{3pt}
\centering
\includegraphics[width=0.9\linewidth]{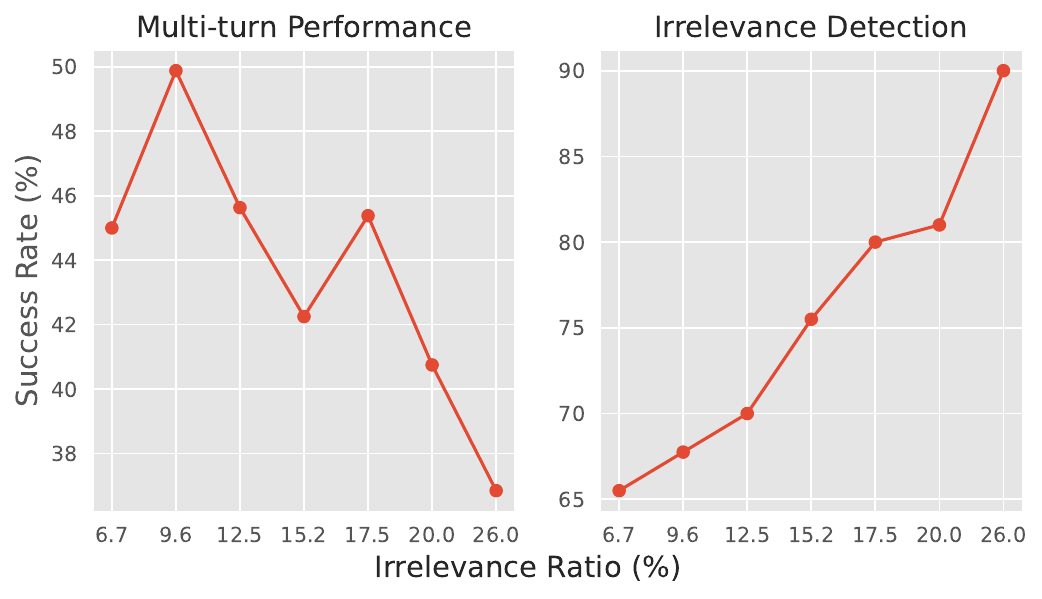}
\caption{The performance when changing the data mixture with different number of irrelevance data.}
\label{fig:irrelevant}
\end{wrapfigure}

We analyze the impact of data mixture to the final performance. As discussed in~\citep{lin2024hammer}, the proportion of queries that involve missing or irrelevant functions would impact the overall behavior of models. We conduct an analysis to study the ratio of single-turn irrelevance samples versus the multi-turn samples. We fix the number of single-turn function call samples and multi-turn samples to 20k and 8k and adjust the ratio of irrelevance samples among 6.7\%, 9.6\%, 12.5\%, 15.2\%, 17.5\%, 20.0\%, and 26.3\%, which corresponds to 2k, 3k $\dots$ 7k and 10k irrelevance samples. We test on our development set which consists of 200 irrelevance test cases and 200 multi-turn test cases. Figure~\ref{fig:irrelevant} exhibits a performance trade-off between multi-turn success and irrelevance detection when adjusting the number of irrelevance examples. The optimal ratio  of irrelevance data that balances the two aspects lies around 15\% to 17\%, based on which we set the final training data mixture in our case. The exact ratio is subject to changes based on different tasks and models but would provide a general guideline when considering data mixture.

\section{Conclusion}
We proposed a novel pipeline, \textsc{Magnet}, for synthesizing multi-turn trajectories for training tool-use LLM agents. Targeted at the challenges in multi-turn FC, We proposed a graph-based multi-turn queries and reference FCs synthesis method to cover those challenges. We further converted those query-reference pairs into trajectories for both SFT and then mDPO training of LLMs. We demonstrated strong performance on agentic benchmarks.

\section*{Acknowledgment}
We thank people from UCLA-NLP and colleagues from Google Cloud AI Research for their valuable feedback during the preparation of the paper.
\bibliographystyle{abbrvnat}
\nobibliography*
\bibliography{custom}

\clearpage
\appendix
\section*{Limitations}
In this section, we discuss the limitations of the work. First, the function signatures we studied in the paper mainly consist of English and pure texts. It is possible some conclusions of this work might not generalize well to other languages and modalities. Future work could consider study multi-lingual and multi-modal tools as an extension to this work.

Second, in our qualitative study, we observe that our trained model might make mistakes when the knowledge retrieved by the tool is conflicted with the internal knowledge of the model. For example, consider a function \texttt{get\_todays\_date}, the tool might return a value that would be changing permanently. However, we found that even with the tool outputs, the model might still output some fixed date such as 2024-05-02. This reflects some limitations in resolving knowledge conflicts within context and internal knowledge.

Third, more exploration abilities could be incorporated into the model in future work. An ideal agent would be able to reflect on their wrong actions and restart the exploration, which is currently limited in our model, due to lack of such data in our training set.
\section{Prompts}
\label{appendix: prompt}
In this section, we list the prompts we used over the data synthesis and model inference process.

\begin{minipage}{16cm}

\noindent {\bf Check nested prompt} We use the following prompt to determine whether two functions are nested:

\texttt{You will be given two function information including their descriptions, parameters, response info etc. Your task is to determine whether the two functions can be nested.
We call two functions to be nested when some parameter values for the later function call can be obtained by the first function call. For example when the first function is convert\_usd\_from\_rmb(rmb\_number=), and the second function is set\_budget\_limit(budget\_limit\_in\_usd=). The two functions are nested because set\_budget\_limit needs a parameter value in dollars and convert\_usd\_from\_rmb could output a dollar value. As another example, when the first function is get\_airport\_symbol\_by\_city(city=,range=), the second function get\_flight\_by\_airport(airport\_symbol=). The two functions are nested because the second function needs a symbol of airport while the first function provides that in the output. Please judge whether the input functions satisfy this nesting relationship. Return two lines: In the first line, If those two functions are nested, output yes, otherwise output no, Use lower case. In the second line, give a brief explanation on why you think they are nested.}

\end{minipage}

\begin{minipage}{16cm}
\noindent {\bf System prompts for training and evaluation} We use the following system prompt following BFCL-v3 for both the training trajectories and the BFCL-v3, ToolQuery inference.

\texttt{You are an expert in composing functions. You are given a question and a set of possible functions. Based on the question, you will need to make one or more function/tool calls to achieve the purpose. If none of the function can be used, point it out. If the given question lacks the parameters required by the function, also point it out. You should only return the function call in tools call sections. If you decide to invoke any of the function(s), you MUST put it in the format of [func\_name1(params\_name1=params\_value1, params\_name2=params\_value2...), func\_name2(params)]. You SHOULD NOT include any other text in the response. 
Here is a list of functions in JSON format that you can invoke.}

For the list of functions, each function is formatted in this way:

\begin{verbatim}
    template = {
        "category": "",
        "tool_name": "",
        "tool_description": "",
        "api_name": "",
        "api_description": "",
        "parameters": {
            "type": "dict",
            "properties": {
            },
            "required": [],
            "optional": [],
        }
    }  
\end{verbatim}



\noindent {\bf Function domain classification prompt} We use the following prompt to classify the domains of functions:

\texttt{You will be given a few domains and a function from one of those domains. You will be given the function name, description, and the required parameters of it. Your task is to classify the function into one of the domains.
The domains are:
`Cybersecurity', `Artificial\_Intelligence', `Commerce', `Advertising', `Payments', `News\_Media', `Cryptography', `Devices', `Business', `eCommerce', `Logistics', `Finance', `Events', `Email', `Business\_Software', `Music', `Database', `Translation', `Jobs', `Gaming', `Monitoring', `func\_source\_code', `Education', `Entertainment', `Visual\_Recognition', `Sports', `SMS', `Media', `Search', `Finance', `Location', `Movies', `Transportation', `Text\_Analysis', `Mapping', `Energy', `Customized', `Medical', `Storage', `Food', `Health', `Video\_Images', ``Science', 'Communication', `Travel', `Social', `Data', `Reward', `Weather'. Return one line with the name of the domain. Or, if you cannot decide on which domain the function belongs to or think the function does not belong to any of the domains, output 'misc'.}
\end{minipage}
\clearpage
\begin{minipage}{16cm}
\noindent {\bf Dependency prompt} We use the following prompt to determine whether any of the candidates function could be neighbors to a target function:

\texttt{You will be given a few API functions. You will also be given a target API. Your task is to create the adjacent list of the target API from those APIs. 
Each element in the adjacent list should be related to the target API.
We say another function is related to the target API if:
1) the output of the target API is the premise of executing the function. For example, the output of fileexists('file.txt') API determines whether we can call downloadfile('file.txt').
2) the output of the target API is exactly the input parameters of the function. For example, when calculating the area of a circle, the function getradius(obj) is the source node and calculate(radius) is the target node.
3) the output of the target API is partial input parameters of the function. For example, when posting something to social media, one might first get the content. In this case, the content = getcontent('file.txt') is the source node and posting(content, id, tags) is the target node.
Notice that the relation might cross the boundary of domains.
For example, when the given APIs are in the domain of weather and travel, it is possible that a weather API could be related to a travel API since the weather determines the travel schedule.
Also, the target API itself should not be in the adjacent list.
For example, if the target API is get\_id, there should not be a get\_id function in the adjacent list.
Return only the adjacency dictionary in a json format. Use exactly the original name of the tool as the key and values. In the adjacency dictionary, the only key is the target API, and each value is a list that contains the relevant APIs for that target API.}

\end{minipage}
\clearpage

\begin{minipage}{16cm}

\noindent {\bf Context distillation for positive trajectories prompt} We use the following prompt for context distillation of positive trajectories:
\texttt{
You are an expert in composing functions. You are given a question and a set of possible functions. Based on the question, you will need to make one or more function/tool calls to achieve the purpose. If none of the function can be used, point it out. If the given question lacks the parameters required by the function, also point it out. You should only return the function call in tools call sections. If you decide to invoke any of the function(s), you MUST put it in the format of [func\_name1(params\_name1=params\_value1, params\_name2=params\_value2...), func\_name2(params)]. You SHOULD NOT include any other text in the response. 
Here is a list of functions in JSON format that you can invoke. Notice that for each question, I already added hint function calls, following the [Hint] key words. Please compose your answer based on those hints while not mentioning those hints explicitly in your responses, i.e., when you decide to invoke function calls, just return the functions, and when you provide textual response, do not mention that there is a hint. Your textual response should summarize the function call outputs. Most of the time the hints are correct answers, just follow it... However, sometimes, those hints might not be perfectly correct, for example, you might see placeholders in the hints parameters like param1=unknow. So, when the hints are not correct, you need to identify them and compose the proper functions by looking for those parameter values from all previous turns. When you see [Hint]: miss function, this means the function needed in this step is missed. You should not simply output miss function in this case but try to use natural language to describe the situation and what functionality is missed. Similarly, when you see [Hint]: missed params, this means that some required parameters for the function is not mentioned in the query, just output some pure texts to ask for the information. However, in your response, do not mention the hint, just answer to the query. When you encounter errors in function outputs, please try composing the functions again based on the error information in the errors. Do not just output textual response at once. **This is important**: when you see the [Hint] contains multiple function calls, i.e., more than one functions should be called for the query, this means those functions are relevant and nested. In this case, at each turn of your response, call only one function. Then, wait for the feedback from the user and then, call the next function. This is because sometimes the parameters of the later functions are missed without the user feedback. For example, when you see [Hint]: func\_name1(params\_name1=params\_value1), func\_name2(params\_name2=params\_value2), you should first output [func\_name1(...)] with the correct parameter values  and wait for the user response. Then, after you get the user response, based on the response, you call the next function [func\_name2(...)] with the correct parameter values.}

\end{minipage}
\clearpage
\begin{minipage}{16cm}

\noindent {\bf Hints selection for negative trajectories} We use the following prompt for the judgement model which is also a Gemini-1.5-pro-002, for determining a negative trajectory hint:
\texttt{You will be given a multi-turn conversation between a user and an agent, the agent response for a single turn, which is possibly a function call, and a reference response. Your task is to judge whether the model response is a correct one based on the reference response. Below are possible error types. When both the reference and the model response are function calls, your judgement is for whether the model response accurately invoke the correct function call.\\
A response might be wrong in the following way:\\
1. Nested function calls: There are missing function calls. Model fails to call some necessary functions because they are not explicitly mentioned in the query.\\
2. Short dependency: There are outputs from a previous function call in this turn that is not used correctly in later function calls.\\
3. Long dependency: There are some parameter values exist in the conversation history but not properly used in this turn.\\
When both the reference and the model response are not function call but general textual response, your judgement is for whether the model response covers all the necessary information but also not hallucination based on the reference response.\\
4. Wrong summarization: whether the model response is a wrong summarization of the reference response.\\
When either one of the reference or the model response is not a function call while the other one is:\\
5. Missed function or parameters: there are some parameter values or functions present or not present in the context while the model thinks the opposite.\\Additional guidelines:
If one of the reference and model responses is function call while the other is not, directly output no.\\
Notice that when you see redundant parameters from the model response when it is function call, it might because it gives all the parameters even the default ones. So, as long as other parameters take the same values, regard this as correct.\\
In the first line, return yes or no. If your answer is no, in the second line, return a number to represent the error type.}

\end{minipage}

\begin{minipage}{16cm}

\noindent {\bf Forth-translation prompt} We use the following prompt for forth-translation to fill in function call parameters to make them executable:
\texttt{Now you are role-playing as a function-calling agent that involves in a multi-turn conversation with a user. You will be given the functions called by the history of this multi-turn conversation, indicated by round numbers. The functions called last round start with [Last Round].You will also be provided with a candidate function in a dictionary format with its descriptions and parameters.
I would like you to generate the function call for the next round using this function signature. Make sure the parameters for this candidate function should be derived from the user query and reference outputs from the last round function call.
Rules:
- You should use the function with the original name without any changes.\\
- For all the functions, make sure your generated function calls contain ALL the required parameters fields from the function documentation. You may also include some optional parameters. However, do not hallucinate any parameters outside of those. Use only the parameters indicated in the required and optional fields of the function documentation.\\
- Then, the parameter values for the new function should be related to the output from last round, please refer to the [Reference Output] for the corresponding values.
- You can have parallel function call with the candidate function, i.e., call the function with different set of parameters, for your new query. However, **do not call more than three parallel functions**.\\
Format:\\Thought:
<the thought on which parameter values to use>\\Answer:
<You need to provide a groundtruth for the function calls that will be invoked in the next round as well as the parameters. Separate your reference function calls by comma. No any other separator is acceptable, only using comma. Also, if any of your parameters are with string value, use double quotation marks to include the parameters. If no answer can be generated, output FINISH in this line>}

\end{minipage}
\clearpage
\begin{minipage}{16cm}

\noindent {\bf Back-translation prompt} We use the following prompt for back-translation from a function signature to a query. The in-context examples are skipped for clarity:

\texttt{Now you are role-playing as a user that involves in a multi-turn conversation with a function-calling agent. You will be given the functions called by the history of this multi-turn conversation, indicated by round numbers. The functions called last round start with [Last Round]. You will also be provided with a list of candidate functions in a dictionary format where the keys are the functions called last round and values are related and candidate functions that can be called in this round. I would like you to generate the query of this round which calls one or multiple functions from the candidate function list. When calling multiple functions, make sure you call no more than three functions at a single round.\\
Rules:\\
- The preferred next round query should be motivated by the outputs from the last round function output. Preferably, those outputs are used as the input parameters for as least one of the functions being called at this round.\\
- You should NOT mention which functions to use in your query explicitly.\\
- After you decide on which function to use, make sure your new query contains information for all the required parameters of the functions you want to call, although some information may be referred to implicitly as the outputs from the last round. If the value for some required parameters are not clear given the context, you may want to create a value for that required parameter but just remember, have information for all required parameters.\\
- Use no parameters besides the parameters indicated in the required and optional fields of the function documentation.\\
- For outputs from the last round, try not to mention the exact parameters that you will use. Instead, use references such as 'the location you just found', 'With the listed items'... to refer to the output of last round that will be leveraged next.\\
- Do not repeat any queries in the conversation history. This means your new query should not call the same function with the same set of parameters as any of the queries in the conversation, even the function exists in the adjacent list.\\
- Avoid using the APIs in [Do not use these APIs].\\
- Try to make the conversation as natural as possible. Mind the logic between two consecutive queries. Do not just create an independent new query.\\
- Below are some examples of good output given conversation history. Please follow the style of conversation and make your new query chained with previous queries.}
\end{minipage}

\section{Training setup}
\label{appendix: training}
We fine-tune Qwen2.5-Coder-7B-instruct and  Qwen2.5-Coder-14B-instruct as the starting point and conduct SFT+RLHF over them. The reason for choosing these base models is that they have been adopted by other strong function calling models as the base model and have demonstrated strong potential for function calling abilities. All experiments are conducted on 16 Nvidia A100 GPUs on the same node. For SFT training, we fine-tune the full parameters for both sizes. We use a fixed max length of 8,172, warm up date of 0.1, Adam~\citep{kingma2014adam} as optimizer and search over learning rate \{1e-5, 5e-5\}, batch size \{64, 128\} with gradient accumulation, and epochs \{1, 2\}. In general, we find that training for 1 epoch works the best. Other parameters are set as in the Section~\ref{exp:setup}. For mDPO, we use LoRA tuning for 14B SFT model with a fixed rank 32 and alpha 64 and fully train the 7B SFT model. We search over learning rate \{5e-7, 1e-6, 5e-6\}, batch size \{32, 64\}, epoch \{1, 2, 3\}, beta \{0.1, 0.01, 0.3\}.

We use the transformer-trl~\footnote{\url{https://github.com/huggingface/trl}} package for training SFT models and use the implementation from~\citet{xiong2024building}, which is also based on transformer-trl, for the mDPO training. 

\section{Data contamination study on BFCL-v3}
\label{appendix: contamimation}
Note that both the ToolQuery test set and the StableToolBench functions have minimal concerns on data contamination. Here, we focus on the BFCL-v3 datasets and study the data contamination of using the backend implementation of python functions of BFCL multi-turn scenarios. We analyze from two aspects: 1) how much performance boosts come from the more general StableToolBench function pool and how much the performance comes from the BFCL-v3 functions; 2) treating the FSP as a sequence of tokens, how much is the n-gram overlap and exact match are there between the test set and training set.

From the first direction, we find that without StableToolBench data, the multi-turn performance on basic, missing function, missing parameters, and long-context are 31.0, 21.0, 17.0, 13.0, with an average of 20.5 score. Comparing to the final performance of 37.5. The general functions from StableToolBench contributes the performance by a large margin.

From the second direction, we observe 0.3\% exact matched underline FSPs on basic multi-turn test cases of BFCL-v3 and 0\% exact match on other categories of multi-turn test cases on BFCL-v3, despite of changed function names. We also examine the 2-gram overlaps on the FSPs between turns, and there are 5.3\% of 2-gram overlaps.
\end{document}